\def\year{2022}\relax
\documentclass[letterpaper]{article} % DO NOT CHANGE THIS
\usepackage{aaai22}  % DO NOT CHANGE THIS
\usepackage{times}  % DO NOT CHANGE THIS
\usepackage{helvet}  % DO NOT CHANGE THIS
\usepackage{courier}  % DO NOT CHANGE THIS
\usepackage[hyphens]{url}  % DO NOT CHANGE THIS
\usepackage{graphicx} % DO NOT CHANGE THIS
\usepackage{natbib}  % DO NOT CHANGE THIS AND DO NOT ADD ANY OPTIONS TO IT
\usepackage{caption} % DO NOT CHANGE THIS AND DO NOT ADD ANY OPTIONS TO IT
\DeclareCaptionStyle{ruled}{labelfont=normalfont,labelsep=colon,strut=off} % DO NOT CHANGE THIS
\usepackage{color}
\usepackage{multirow}
\usepackage{subfigure}
\usepackage[linesnumbered, ruled]{algorithm2e}
\usepackage{xspace}
\usepackage{dsfont}
\usepackage{epsfig}
\usepackage{epstopdf}
\usepackage{booktabs}
\usepackage{amsmath}
\usepackage{amsthm}
\usepackage{amsfonts}
\usepackage{bm}

\newtheorem{mydef}{Definition}

\newcommand{\paratitle}[1]{\vspace{1.5ex}\noindent\textbf{#1}}
\newcommand{\ie}{\emph{i.e.,}\xspace}

\newcommand{\eg}{\emph{e.g.,}\xspace}

\newcommand{\uem}[1]{\emph{\underline{#1}}}

\newcommand{\name}{STDEN\xspace}
\newcommand{\figureautorefname}{Fig.}
\newcommand{\tableautorefname}{Tab.}
\DeclareMathOperator{\udiv}{div}

\DeclareMathOperator{\wh}{where}

\title{\name: Towards Physics-Guided Neural Networks for Traffic Flow Prediction}

\author {
    Jiahao Ji, \textsuperscript{\rm 1}
    Jingyuan Wang, \textsuperscript{\rm 1,\rm 2}\thanks{Corresponding author: jywang@buaa.edu.cn}
    Zhe Jiang, \textsuperscript{\rm 3}
    Jiawei Jiang, \textsuperscript{\rm 4}
    Hu Zhang \textsuperscript{\rm 4}
}
\affiliations {
    \textsuperscript{\rm 1} State Key Laboratory of Software Development Environment,\\ School of Computer Science \& Engineering, Beihang University, Beijing, China\\
    \textsuperscript{\rm 2} Peng Cheng Laboratory, Shenzhen, China\\
    \textsuperscript{\rm 3} Department of Computer \& Information Science \& Engineering, The University of Florida\\
    \textsuperscript{\rm 4} MOE Engineering Research Center of Advanced Computer Application Technology,\\School of Computer Science \& Engineering, Beihang University, Beijing, China\\
    \{jiahaoji, jywang, jwjiang, zhanghu2006\}@buaa.edu.cn, zhe.jiang@ufl.edu
}

\begin{document}

\maketitle

\begin{abstract}
  High-performance traffic flow prediction model designing, a core technology of Intelligent Transportation System, is a long-standing but still challenging task for industrial and academic communities. The lack of integration between physical principles and data-driven models is an important reason for limiting the development of this field. In the literature, \emph{physics-based} methods can usually provide a clear interpretation of the dynamic process of traffic flow systems but are with limited accuracy, while \emph{data-driven} methods, especially deep learning with black-box structures, can achieve improved performance but can not be fully trusted due to lack of a reasonable physical basis. To bridge the gap between purely data-driven and physics-driven approaches, we propose a physics-guided deep learning model named Spatio-Temporal Differential Equation Network (\name), which casts the physical mechanism of traffic flow dynamics into a deep neural network framework. Specifically, we assume the traffic flow on road networks is driven by a latent potential energy field (like water flows are driven by the gravity field), and model the spatio-temporal dynamic process of the potential energy field as a differential equation network. \name absorbs both the performance advantage of data-driven models and the interpretability of physics-based models, so is named a \emph{physics-guided} prediction model. Experiments on three real-world traffic datasets in Beijing show that our model outperforms state-of-the-art baselines by a significant margin. A case study further verifies that \name can capture the mechanism of urban traffic and generate accurate predictions with physical meaning. The proposed framework of differential equation network modeling may also cast light on other similar applications.
\end{abstract}

\section{Introduction}\label{sec:intro}

Rapid urbanization has brought about the growth of urban population, and presented huge transportation and sustainability challenges to modern cities. Intelligent Transportation System (ITS) has become an active research area because of its potential to improve transportation efficiency and solve the sustainability problem of cities~\cite{snyder2019streets}. As the core technology of the ITS, {\em traffic flow prediction}, aiming at forecasting the future status of traffic systems given historical observations, plays a crucial role in many important urban applications, such as public safety, congestion management and navigation~\cite{li2020autost}. %On the other hand, traffic flow prediction is also a challenge because traffic data in real-world road networks are complex, noisy, and with large fluctuations.

In the literature, traffic flow prediction methods mainly fall into two categories: \emph{physics-based} and \emph{data-driven}. The former one usually relies on traffic flow theory~\cite{ni2015traffic}, which represents traffic system as coupled Differential Equations (DEs). Traffic flow prediction is then achieved through conducting system simulation governed by these DEs. The physics-based models are able to guarantee the simulation results consistently represent the traffic dynamics over the entire domain, not only where it was calibrated by observation data. However, these models usually make strong assumptions about traffic flow with a small set of parameters~\cite{mo2020physics}, which may not be able to capture the complex human behaviors and the uncertain factors in real-world traffic. Besides, the simulation process relies on solving DEs through numerical differentiation and integration techniques, which requires a lot of computing resources~\cite{wang2020towards}.

% networks~\cite{li2018diffusion} and temporal convolutional networks~\cite{wu2019graph}.
% rnn: bai2020adaptive, tcn: li2021spatial

The second category is data-driven methods, which usually utilize historical observational data to train a statistical learning model, and then use the trained model to generate predictions. Among the data-driven methods, the most representative branch is traffic flow prediction based on deep learning. For example, using recurrent neural networks~\cite{li2018diffusion} or temporal convolution~\cite{wu2019graph} to model temporal dependencies, using convolutional neural networks~\cite{tang2020joint} to capture spatial correlations, and using graph convolutional~\cite{song2020spatial, tian2021spatial} to introduce road network information into traffic prediction. In recent years, with a huge volume of traffic data becoming available, the deep learning-based data-driven methods have drawn great attention from both industry and academia, and achieved great success in many real-world applications. However, these methods also have defects. First, without the physical knowledge to guarantee generalization ability, the data-driven models are very likely to lose effectiveness in scenarios that are not sampled by the training data. Second, the ``black-box'' structure of deep learning models introduces unknown risks in the ITS, which may cause potential threats to urban safety.

To bridge the gap between data-driven and physics-based approaches, we raise a hybrid modeling paradigm, {\em~Physics-Guided Deep Learning}, for traffic flow prediction. Specifically, we propose a Spatio-Temporal Differential Equation Network (\name\footnote{The code is available at \url{https://github.com/Echo-Ji/STDEN}}) that combines the physical mechanism of traffic dynamics and end-to-end deep learning into a whole framework. Our idea is based on the key assumption that traffic flow on road networks is driven by a latent potential energy field (like water flows are driven by the gravity field). The latent potential energy field follows physical constraints that reflect the transport of energy~\cite{lienhard2008doe}. These constraints take the form of differential equations (DEs) commonly used in physics. To capture complex functions in the potential energy field DE in a learnable manner, we extend the existing ordinary DE network~\cite{chen2018neural} by replacing the differential operator with graph Laplacian and design a graph neural network for spatio-temporal road networks. The overall framework of \name consists of an encoder that maps traffic flow into latent potential energy fields, a DE network that predicts the dynamics of potential energy fields continuously over time, and a decoder that generates traffic flow predictions from the latent potential energy fields. Evaluations on real-world traffic datasets show that \name consistently outperforms state-of-the-art traffic prediction baselines by a large margin in terms of prediction accuracy. Moreover, the learned potential energy field can potentially reveal the evolution of urban dynamics, thereby explaining changes in traffic flow. In summary, our contribution is three-fold:
\\$\bullet$ We model the fundamental physical mechanism of urban dynamics using potential energy fields, and introduce the physical mechanism into a data-driven deep learning model for traffic prediction. To the best of our knowledge, this is the first work that proposes a physics-based and data-driven mixed, \ie physics-guided, deep learning model for traffic flow prediction on road networks.
%$\bullet$ We study the physics-guided deep learning mode for traffic flow prediction and model the physical mechanism of urban dynamics using potential energy fields. To the best of our knowledge, this is the first work that proposes a data-driven and physics-driven mixed, \ie, physics-guided, traffic flow prediction.
%We propose a data-driven and physics-driven mixed, \ie physics-guided, deep traffic flow prediction model, where the concept of potential energy fields is proposed to model dynamic of urban traffic and, where the physical mechanism of urban traffic dynamics is modeled as potential energy fields In this study, we model the physical process of urban dynamics using potential energy fields, and mix the physical concept of potential energy fields into deep learning models which gives a fundamental explanation of the dynamic of urban traffic. To the best of our knowledge, this is the first work to introduce the potential energy field concepts into deep learning for road network traffic flow prediction. %that proposes a data-driven and physics-driven mixed, \ie physics-guided, traffic flow prediction. %We study the physics-guided deep learning methods for traffic flow prediction and
\\$\bullet$ We propose a novel hybrid deep learning model, \name, that unifies the traffic potential energy field DE and neural networks into one framework. This modeling paradigm may cast light on other DE-based applications, such as weather forecasting and epidemic prediction.
%The model describes the spatial and temporal dependencies simultaneously in an ODE, thus we utilize neural ODE techniques to model the continuous dynamics.
\\$\bullet$ We conduct extensive experiments on three real-world traffic datasets, and the proposed method achieves significant improvement over state-of-the-art baselines. Moreover, a case study confirms that the learned potential energy field can reveal the physical mechanism of traffic flow and provide interpretability for the deep traffic prediction model.

\section{Preliminaries}\label{sec:pre}

We study the traffic flow prediction problem over urban road networks. A list of major symbols is in \tableautorefname~\ref{tab:sym&desc}.

\begin{mydef}[Road Network]
A \emph{Road Network} is a directed graph $\mathcal{G} = (\mathcal{V}, \mathcal{E}, {W})$, where $\mathcal{V} = \{v_1, \dots, v_n\}$ is a set of nodes, $\mathcal{E} \subseteq \mathcal{V} \times \mathcal{V}$ is a set of edges, and ${W} \in \mathbb{R}^{n \times n}$ is a weighted adjacency matrix. A node $v_i \in \mathcal{V}$ represents a road junction or a road end, while an edge $e_{ij} \in \mathcal{E}$ represents a directed road segment from node $v_i$ to node $v_j$.
\end{mydef}

\begin{mydef}[Traffic Flow]
We define \emph{Traffic Flow} as a feature of edges in the road network. Given an edge $e_{ij}$, the overall traffic flow during a give time period is denoted as $f_{ij}$. We express the traffic flow of the whole road network as a vector $\bm{f} = ({f}_{ij}) \in \mathbb{R}^{|\mathcal{E}|}$. $|\mathcal{E}|$ is the size of the edge set.
\end{mydef}

\begin{table}[t]\small
  \centering
  \begin{tabular}{lll}
  \toprule
    Sym.   &  Domain & Descriptions \\
  \midrule
      % $\mathcal{G}(\mathcal{V}, \mathcal{E}, \bm{W})$ & - & Road network \\
      % $\nabla,\udiv,\Delta$ & - & Gradient, divergence, Laplacian operator\\
      $n$ & $\mathbb{R}$ & Number of nodes in a road network \\
      ${W}$ & $\mathbb{R}^{n \times n}$ & Adjacency matrix of a road network \\
      $\bm{f}^{(t)}$ & $\mathbb{R}^{\vert\mathcal{E}\vert}$  & Traffic flow of a road network at time $t$\\
      % $\textbf{F}^{(1:t)}$ & $\mathbb{R}^{t\times e \times 1}$ & Traffic flow field sequence. \\
      $\bm{z}^{(t)}$ & $\mathbb{R}^{n}$  & Potential field of a road network at time $t$ \\
      % $T$ & $\mathbb{R}$ & Historical sequence length\\
      % $H$ & $\mathbb{R}$ & Prediction horizon\\
      $\bm{u}, \bm{q}$ & $\mathbb{R}$ & Energy density, energy flux\\
      % $$ & $\mathbb{R}$ & \\
      $\bm{\phi}$ & $\mathbb{R}^{n}$ & Node volume\\
      $\alpha^{-1}$ & $\mathbb{R}$ & Contribution ratio of traffic flow\\
  \bottomrule
  \end{tabular}
  \caption{List of major symbols and descriptions.}\label{tab:sym&desc}
\end{table}

\begin{figure}[t]
  \centering
  \includegraphics[width=0.98\columnwidth]{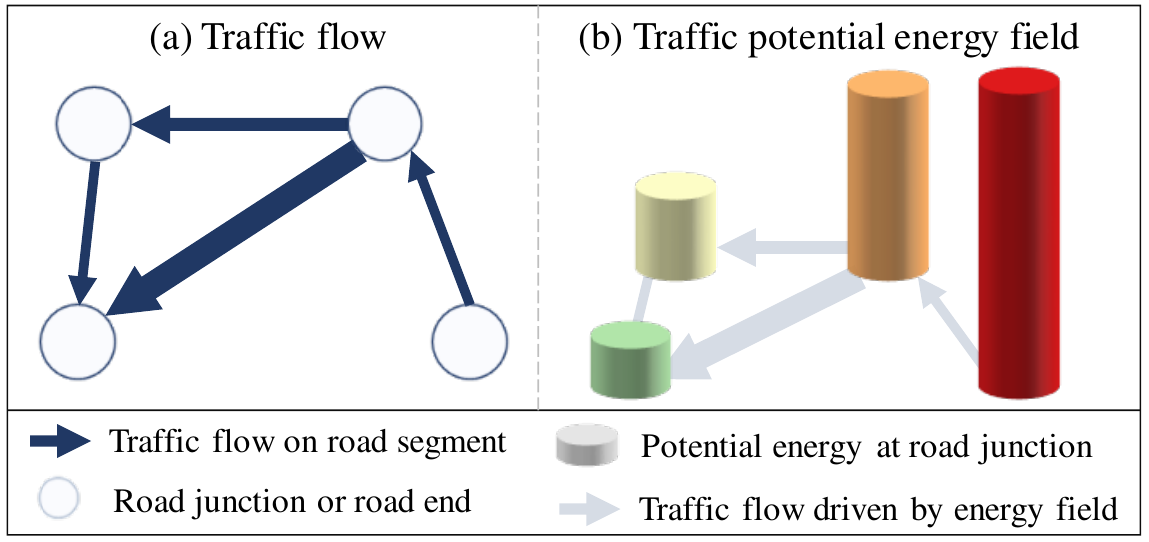}
  \caption{Illustration of traffic flow and potential energy field defined on road networks. The arrow indicates the traffic flow direction while its size denotes the flow volume. In panel (b), higher cylinder means more potential energy, and the traffic flow volume (arrow size) increases with the energy gradients between adjacent nodes.}
  \label{fig:flow_potential}\vspace{-.3cm}
\end{figure}

% We assume the traffic flows in a road network are driven by a latent potential energy field, which is defined as follows.

\begin{mydef}[Traffic Potential Energy Field (PEF)]~\label{def:PEF}
For each node $i$ of a road network $\mathcal{G}$, we define the \emph{Traffic Potential Energy} $z_i \in\mathbb{R}$ as its feature. The \emph{Traffic Potential Energy Field} on the whole road network is denoted as $\bm{z} = (z_{1}, \dots, z_n)^{\top} \in \mathbb{R}^{n}$.
\end{mydef}

As shown in \figureautorefname{~\ref{fig:flow_potential}}, the potential energy field on the road network is the latent dominated force of the traffic flow, which is similar to the gravity field driving water flow. The traffic flow of the corresponding edge is the potential energy gradient between adjacent nodes,
\begin{equation}\label{eq:energy2flow}\small
  {f}_{ij} = - (\nabla z)_{ij} = - (z_i - z_j),
\end{equation}
where $\nabla$ is the graph gradient operator, and the negative sign ahead of $\nabla$ indicates the direction of flow. According to Eq.~\eqref{eq:energy2flow}, the traffic flow on the road network can be explained as a ``energy transport'' process of the traffic potential energy field. This is a fundamental insight of our model.

\begin{mydef}[Flow-sequence and PEF-sequence]
We divide time as regular time slices. Let {\small $\bm{f}^{(t)}$} represent traffic flow observed at time slice $t \in \mathbb{N}$. \emph{Flow-sequence} is the time series of traffic flow {\small $\bm{F}^{(0:t)} = \left(\bm{f}^{(0)}, \dots, \bm{f}^{(t)}\right)$}, while \emph{PEF-sequence} is that of PEF {\small $\bm{Z}^{(0:t)} = \left(\bm{z}^{(0)}, \dots, \bm{z}^{(t)}\right)$}.
\end{mydef}

\subsection{Problem Definition}\label{subsec:problem-def}

Based on the basic concepts above, we formally define the problem of traffic flow prediction as follows.
\\$\bullet$ \textbf{Input:} The history flow-sequence from time {\small $t-T+1$} to {\small $t$}, {\small $\bm{F}^{(t-T+1: t)}$}, the future flow-sequence from time {\small $t+1$} to {\small $t+H$}, {\small $\bm{F}^{(t+1: t+H)}$}, and the corresponding road network $\mathcal{G}$.
\\$\bullet$ \textbf{Output:} A model $m(\cdot)$ satisfying {\small $\hat{\bm{F}}^{(t+1: t+H)} = m\left(\bm{F}^{(t-T+1: t)}\right)$}.
\\$\bullet$ \textbf{Objective:} Minimizing prediction errors of training data.%, where the prediction errors are measured using Mean Absolute Error as
%\begin{equation}\label{}\small
%    \mathrm{MAE} = \frac{1}{H}\sum_{i=t+1}^{t+H} \left|\hat{\bm{f}}^{(i)} - {\bm{f}}^{(i)}\right|.
%\end{equation}

\section{Physics-Guided Traffic Flow Modeling}\label{sec:method}

\subsection{The Model Framework}

\begin{figure}[t]
  \centering
  \includegraphics[width=1.0\columnwidth]{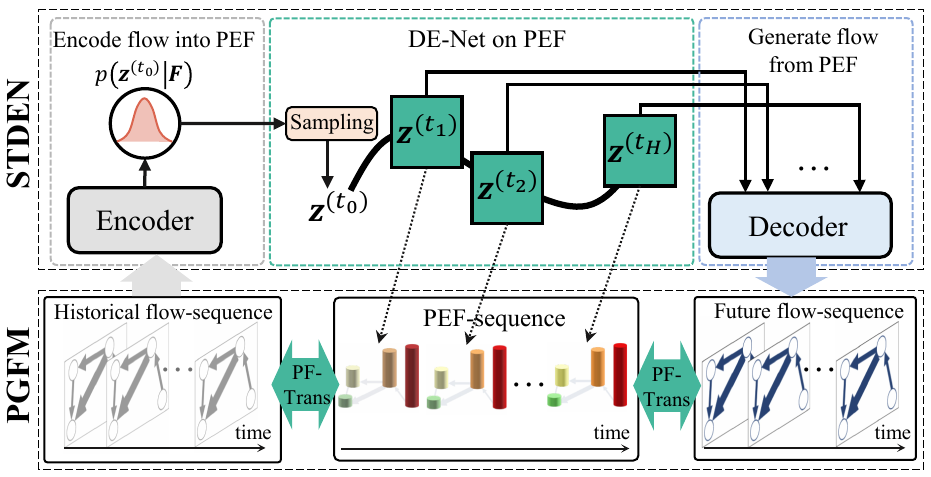}
  \caption{Overview of the proposed physics-guided model. \name: Spatio-Temporal Differential Equation Network. PGFM: Physics-Guided traffic Flow Modeling. PF-Trans: transformation between the PEF and traffic flow (Eq.~\eqref{eq:potn_flow}).}
  \label{fig:framework}%\vspace{-.4cm}
\end{figure}

Data-driven methods directly model the correlation between {\small $\bm{F}^{(t-T+1: t)}$} and {\small $\bm{F}^{(t+1: t+H)}$} to generate predictions. We innovatively propose a physics-guided traffic flow prediction framework (see \figureautorefname{~\ref{fig:framework}}), consisting of two parts: an explicit deep learning model \name and an implicit physical dynamic process PGFM. In PGFM, \ie {\em Physics-Guided traffic Flow Modeling}, we use a physics-based continuity equation to model the transformation between the PEF and the traffic flow (denoted as PF-Trans in \figureautorefname{~\ref{fig:framework}}). Details of the implicit PGFM will be given in this section.

Over the implicit physical dynamic process, we implement a deep neural network model, {\em~Spatio-Temporal Differential Equation Network} (\name), which contains three components: an RNN-based network to encode the traffic flow-sequence as the initial state of PEF, a DE network to predict the evolution of the PEF, and a decoder governed by Eq.~\eqref{eq:energy2flow} to generate traffic flow from the predicted PEF. Note the layers of DE-Net correspond to the dynamic of potential energy in PGFM. Details of \name are in the next section.

In our model, the physical dynamic process and the deep learning model are unified under the same framework, that is why we call our model a {\em ``physics-guided''} neural network.

\subsection{Physics-Based Continuity Equation}

In this section, we use a continuity equation to model the relations between the flow-sequence and PEF-sequence. First, we provide a brief introduction to the continuity equation.

In physics, the continuity equation describes the transport of some physical quantity, such as mass, energy, momentum, electric charge~\cite{lienhard2008doe}. Here we take the transport of energy as an example. Using the continuity equation, the energy transport process over a set of locations could be described as
\begin{equation}\label{eq:ctnu}\small
    \frac{\partial \bm{u}}{\partial t} + \udiv \bm{q} = 0,
\end{equation}
where $\bm{u}$ is the energy density (energy per unit volume), $\bm{q}$ is the {\em Energy Flux} (a measure of energy ``transport''), and $\udiv$ is the divergence operator. 
% For a 3D Cartesian coordinates, the term $\udiv \bm{q}$ can be expressed as
% \begin{equation}\small%\nonumber
%     \udiv \bm{q} = \left(\frac{\partial}{\partial x_1}, \frac{\partial}{\partial x_2}, \frac{\partial}{\partial x_3}\right) \cdot \left(q_{x_1}, q_{x_2}, q_{x_3}\right) = \frac{\partial q_{x_1}}{\partial {x_1}} + \frac{\partial q_{x_2}}{\partial {x_2}} + \frac{\partial q_{x_3}}{\partial {x_3}},
% \end{equation}
The differential equation in Eq.~\eqref{eq:ctnu} gives a relation between the volume of the energy and the ``transport'' of that energy, \ie the change of energy density leads to energy transport in the space. For example, heat always transports from hot locations to cold locations.

%According to Eq.~\eqref{eq:energy2flow}, we express traffic flow on the road network as an energy transport among a Potential Energy Field, therefore, we could use a continuity equation to model the relations between flow-sequence and PEF-sequence of a road network.

\subsection{Continuity Equation for Traffic PEF}

The traffic flow driven by traffic potential energy fields in a road network is naturally a specific example of the continuity equation. In \tableautorefname{~\ref{tab:analogy}}, we draw an analogy between the potential energy-driven traffic flow and the energy continuity equation, where traffic flow on a road network could be considered as energy transport within a Potential Energy Field. In this section, we propose a continuity equation for traffic PEF to describe the dynamic relations between flow-sequence and PEF-sequence of a road network.

\begin{table}[t]
    \centering
    \begin{tabular}{|rcl|}
    \hline
    Potential energy field & $\longrightarrow$ & Energy field \\
    Potential energy per unit volume  & $\longrightarrow$ & Energy density \\
    Traffic flow & $\longrightarrow$ & Energy flux \\
    \hline
    \end{tabular}
    \caption{Analogy from potential energy-driven traffic flow to energy continuity equation.}\label{tab:analogy}\vspace{-.3cm}
\end{table}

\paratitle{Energy Density and Traffic Potential Energy.} Given a road network $\mathcal{G} = \{\mathcal{V}, \mathcal{E}, W\}$, for $v_i \in \mathcal{V}$, we define it having a {\em Energy Density} $u_i$. The potential energy of the $v_i$ is proportional to its energy density as $z_i = \phi_i \cdot u_i$, where $\phi_i$ is the trainable node volume. The node volume is determined by the endogenous feature of the node, which is in analogy with the mass in the gravity field. For all nodes in $\mathcal{V}$, we have the relation between the potential energy {\small $\bm{z} = (z_{1}, \dots, z_n)^{\top}$} and the energy density {\small $\bm{u} = (u_{1}, \dots, u_n)^{\top}$}
\begin{equation}\label{eq:potn2dens}\small
    \bm{z} = \bm{\phi} \odot \bm{u},
\end{equation}
where {\small $\bm{\phi} = (\phi_1, \dots, \phi_n)^{\top}$} and {\small $\odot$} is the Hadamard product.
% \ie the element-wise product.

\paratitle{Energy Flux and Traffic Flow.} In a traffic system defined on $\mathcal{G}$, the traffic potential energy can only be transported along the edges (road segments) in the edge set $\mathcal{E}$. 
% Therefore, the {\em Energy Flux} in the traffic PEF is defined along the edges. For $e_{ij} \in \mathcal{E}$, we define its {\em Energy Flux} as ${q}_{ij}$. 
So we define the {\em Energy Flux} of $e_{ij} \in \mathcal{E}$ as ${q}_{ij}$. 
We consider the traffic flow ${f}_{ij}$ as the major component of the energy flux ${q}_{ij}$, and introduce a shared parameter $\alpha$ to measure the contribution ratio of the traffic flow to energy flux for every edge,
%  as ${q}_{ij} = \alpha {f}_{ij}$. For the whole road network, we have
\begin{equation}\label{eq:flow2flux}\small
    \bm{f} = \alpha^{-1} \bm{q},
\end{equation}
where {\small $\bm{f} = ({f}_{ij})^{\top}$} and {\small $\bm{q} = ({q}_{ij})^{\top}$}.

\paratitle{Potential Energy DE Function.} Plugging Eq.~\eqref{eq:potn2dens} and \eqref{eq:flow2flux} into the energy continuity equation in Eq.~\eqref{eq:ctnu}, we have
\begin{equation}\label{eq:potn_flow}\small
    \bm{\phi}^{-1} \odot \frac{\partial \bm{z}}{\partial t} + \alpha \udiv \bm{f} = 0,
\end{equation}
which is the continuity equation for energy transport in the traffic potential energy field. According to Eq.~\eqref{eq:energy2flow}, traffic flow on edges of a road network is the gradient of potential energy between adjacent nodes, \ie {\small $\bm{f} = - \nabla \bm{z}$}. Using it to replace the traffic flow $\bm{f}$ in Eq.~\eqref{eq:potn_flow}, we have
\begin{equation}\label{eq:potn_org}\small
     \bm{\phi}^{-1} \odot \frac{\partial \bm{z}}{\partial t} - \alpha \udiv \nabla \bm{z} = 0.
\end{equation}
This function is a {\em Differential Equation} (DE) about the potential energy field $\bm{z}$, so we name it as {\em Potential Energy Field DE}.
The graph Laplacian operator $\Delta$ is given by the divergence of the gradient~\cite{bronstein2017geometric}, \ie {\small $\Delta = - \udiv \nabla$}, so the potential energy field DE can be transformed into
% so the potential energy DE can be expressed as the form of
\begin{equation}\label{eq:potn_fin}\small
    \frac{\partial \bm{z}}{\partial t} = - \bm{\phi} \odot \left(\alpha \Delta \bm{z}\right).
\end{equation}

\paratitle{Physics Interpretation of the Potential Energy Field DE.} Eq.~\eqref{eq:potn_fin} is the dynamical equation to express the evolution of the PEF-sequence {\small $\bm{Z} = \left(\bm{z}^{(t)}\right)$}. From a discrete perspective, Eq.~\eqref{eq:potn_fin} could be expressed as the form of
\begin{equation}\small\label{eq:potn_fin_d}
    \bm{z}^{(t+1)} = \bm{z}^{(t)} - \bm{\phi} \odot \left(\alpha \Delta \bm{z}^{(t)}\right),
\end{equation}
which gives an evolution trajectory of the PEF-sequence {\small $\bm{Z}$}.

%In Eq.~\eqref{eq:potn_fin}, the change of potential energy $\bm{z}$, \ie ${\partial \bm{z}}/{\partial t}$, depends on $\Delta \bm{z}$. The graph Laplacian operator $\Delta$ gives the difference between the potential energy $z_i$ of a node and the average potential energy of the node $i$'s neighbors, indicating the process of traffic potential energy transports from high potential energy nodes to their low potential energy neighbors.

%Therefore, Eq.~\eqref{eq:potn_fin} expresses that the potential energy transports from high potential energy nodes to their low potential energy neighbors.

%This process corresponds to urban traffic flows from high traffic potential energy road junctions to low traffic potential energy  road junctions.

\section{Spatio-Temporal Differential Equation Network}\label{sec:ODE}

In Eq.~\eqref{eq:potn_fin}, the evolution of PEF-sequence is expressed as a Potential Energy Field  DE form. In this section, we implement Eq.~\eqref{eq:potn_fin} using a neural network approach to enhance the modeling capability of the potential energy DE for real-world traffic data. The framework of our model is in \figureautorefname{~\ref{fig:framework}}.

\subsection{Differential Equation Network}

%We first give an introduction of the differential equation approach.

Our method is based on the neural ordinary {\em Differential Equation Network} (DE-Net)~\cite{chen2018neural}, which is a kind of model that generalizes standard layer-to-layer neural networks as a continuous form. Specifically, in standard residual networks~\cite{he2016deep}, the layer-to-layer state update process is in the form of
\begin{equation}\label{eq:resnet}\small
    \bm{h}_{t+1} = \bm{h}_t + \mathcal{F}(\bm{h}_t, \theta_{t}),
\end{equation}
where {\small $\bm{h}_t$} is hidden state of (\ie layer outputs) of the {\small $t$}-th layer\footnote{In our model, the layer of DE-Net just corresponds to time slices, so we use $t$ as the index of the network layers in Eq.~\eqref{eq:resnet}.}. {\small $\mathcal{F}(\cdot)$} is the repeated network structure of the residual networks and {\small $\theta_{t}$} is the network parameters at layer {\small $t$}.

Eq.~\eqref{eq:resnet} could be rewritten as a generalized form as
\begin{equation}\label{eq:div_resnet}\small %\nonumber
    \frac{\bm{h}_{t+l} - \bm{h}_t}{(t + l) - t} = \mathcal{F}(\bm{h}_t, \theta_t),
\end{equation}
where $l$ is the step size and its value is 1 in residual networks. When $l\rightarrow 0$, the update process of state $\bm{h}$ becomes
\begin{equation}\label{eq:cont_resnet}\small %\nonumber
    \frac{\partial \bm{h}(t)}{\partial t} = \mathcal{F}\left(\bm{h}(t), t, \theta\right),
\end{equation}
which is a continuous version of the residual networks. In Eq.~\eqref{eq:cont_resnet}, the deep residual neural network becomes a dynamic system that is governed by an ordinary differential equation~\cite{ruthotto2019deep, lu2018beyond}. Moreover, the function {\small $\mathcal{F}(\cdot, \theta)$} is a neural network, which provides powerful representation learning capacity for complex data.
%  weinan2017proposal

\subsection{DE-Net for Traffic Potential Energy Fields}

\paratitle{DE-Net on Potential Energy Fields.} Inspired by the neural ordinary differential equation network, we model the potential energy field DE in Eq.~\eqref{eq:potn_fin} as,
\begin{equation}\label{eq:ST-ODE}\small
    \frac{\partial{\bm{z}^{(t)}}}{\partial{t}} = \mathcal{F_G}\left(\Phi, t, \bm{z}^{(t)}\right),
\end{equation}
where $\Phi$ represents all trainable parameters including $\bm{\phi}$ and $\alpha$. The function $\mathcal{F_G}$ is a neural network guided by the physics model in Eq.~\eqref{eq:potn_fin}.
According to Eq.~\eqref{eq:potn_fin}, we express the function $\mathcal{F_G}$ as a residual graph convolution network (GCN) form, where the repeated neural network layer is in the form of
\begin{equation}~\label{eq:gcn}\small
    \mathcal{F_G}\left(\Phi, t, \bm{z}^{(t)}\right) = - \bm{\phi} \odot \mathrm{Tanh}\left(\alpha \Delta \bm{z}\right).
\end{equation}
where $\Delta$ is the graph Laplacian operator to calculate the differences between the state $z_i$ of the node $i$ and its neighbor's. This is equivalent to using $\alpha$ as a convolution kernel to aggregate the states of nodes in a receptive field. $\mathrm{Tanh}(\cdot)$ is a Hyperbolic Tangent activation function. The results of the convolution are combined using the weights from $\phi_i$.

From the spatial perspective, if we set $t$ as a discrete value, Eq.~\eqref{eq:gcn} is equivalent to a residual GCN, where inputs of each layer are ${z}^{(t)}$ for all nodes of the road network. From the temporal perspective, the time $t$ is continuous, meaning we can calculate $\bm{z}^{(t)}$ for any $t\in \mathbb{R}$. Therefore, we name the proposed model based on the differential equation network in Eq.~\eqref{eq:gcn} as {\em Spatio-Temporal DE Network}.

% Eq.~\eqref{eq:gcn} could also be considered as a variant of recurrent neural network because it can be expressed as
% \begin{equation}\label{}\small
%   \bm{z}^{(t+1)} = \mathcal{F_G}\left(\Phi, t, \bm{z}^{(t)}\right) = \bm{z}^{(t)} + \frac{\partial{\bm{z}^{(t)}}}{\partial{t}}.
% \end{equation}
% Therefore, we name the proposed model based on the neural ordinary differential equation in Eq.~\eqref{eq:gcn} as a {\em Spatio-Temporal Neural ODE} model. Moreover, for our neural ODE model, the time $t$ is continuous, meaning we can calculate $\bm{z}^{(t)}$ for any $t\in \mathbb{R}$. This improved the modeling capability of our model~\cite{queiruga2020continuous}.

\paratitle{Encode Traffic Flow into Potential Energy Fields.} In our model, the PEF-sequence $\bm{Z}^{(t_1:t_H)}$ can be calculated using a neural ODE solver~\cite{chen2018neural} for a given initial state $\bm{z}^{(t_0)}$,
% \begin{equation}\label{eq:traj-z}\small
% \begin{split}
%     \bm{Z}^{(t_1:t_H)} &= \mathrm{ODEsolver}\left(\mathcal{F_G}, \Phi, \bm{z}^{(t_0)}, [t_0, \dots, t_H]\right), \\
%     &\quad \wh \quad \bm{z}^{(t_0)} \sim p\left(\bm{z}^{(t_0)}\right),
% \end{split}
% \end{equation}
\begin{equation}\label{eq:traj-z}\small
  \bm{Z}^{(t_1:t_H)} = \mathrm{ODEsolver}\left(\mathcal{F_G}, \Phi, \bm{z}^{(t_0)}, [t_0, \dots, t_H]\right),
  \end{equation}
where $\bm{z}^{(t_0)}$ is sampled from a distribution $p\left(\bm{z}^{(t_0)}\right)$. We implement the distribution as a conditional probability distribution of historical traffic flow sequence $\bm{F}^{(t_0-T+1:t_0)}$. Specifically, we let $\bm{z}^{(t_0)}$ is generated from a Gaussian distribution, where the mean and standard deviation are determined by the historical traffic flow sequence $\bm{F}^{(t_0-T+1:t_0)}$ as
\begin{equation}\label{eq:init-z0}\small
\begin{split}
    p\left(\bm{z}^{(t_0)} \middle| \bm{F}^{(t_0-T+1:t_0)}\right) &= \mathcal{N}\left(\mu_{\bm{z}^{(t_0)}}, \sigma_{\bm{z}^{(t_0)}}\right), \\
    \wh \; \left\{\mu_{\bm{z}^{(t_0)}}, \sigma_{\bm{z}^{(t_0)}}\right\} &= g\left(\mathrm{GRU}\left(\bm{F}^{(t_0-T+1:t_0)}\right)\right).
\end{split}
\end{equation}
We employ Gated Recurrent Unit (GRU)~\cite{chung2014empirical} as encoder to extracts information from $\bm{F}^{(t_0-T+1:t_0)}$. 
% We employ Gated Recurrent Unit (GRU)~\cite{chung2014empirical} to implement $\mathrm{RNN}(\cdot)$. 
$g(\cdot)$ is a fully connected network to translate the final hidden states of GRU into the mean and standard deviation of $\bm{z}^{(t_0)}$.

To achieve a differentiable ``sampling'' operation, we adopt a reparametrization trick~\cite{kingma2014auto} to implement generation process $g(\cdot)$ of $\bm{z}^{(t_0)}$ in Eq.~\eqref{eq:init-z0}. In specific, given a batch of training data, we calculate {\small $z_i^{(t_0)}$} of each node $i$ as
\begin{equation}\label{eq:reparam}\small
    z_i^{(t_0)} = \mu_{\bm{z}^{(t_0)}} + \epsilon_i \sigma_{\bm{z}^{(t_0)}},
\end{equation}
where $\epsilon_i$ is sampled from a standard normal distribution {\small $\mathcal{N}(0,1)$}. In this way, $z_i^{(t_0)}$ for a given batch of training data are fixed, and therefore, Eq.~\eqref{eq:reparam} is differentiable in backpropagation algorithm for neural network training. In the prediction phase, the $\epsilon_i$ is sampled for every input example.

\paratitle{Generate Traffic Flow from Potential Energy Fields.} Given the initial $\bm{z}^{(t_0)}$, we can calculate the entire PEF-sequence $\bm{Z}^{(t_1:t_H)}$ for future time steps. Next, we generate the flow-sequence $\bm{F}^{(t_1:t_H)}$ from $\bm{Z}^{(t_1:t_H)}$ by Eq.~\eqref{eq:energy2flow}.
% as
% \begin{equation}\small
%     \hat{\bm{f}}^{(t+h)} = r\left(\bm{z}^{(t_h)}\right), 0 < h < H, h \in \mathbb{N},
% \end{equation}
% where $r(\cdot)$ is a fully connected layer. The parameters of $r(\cdot)$ are shared by all nodes $\mathcal{V}$ and time-invariant along $t$.

\subsection{The Model Training}

With the pipeline introduced above, we build a physics-guided latent variable model entitled {\em Spatio-Temporal Differential Equation Network} (\name). Using the backpropagation algorithm, the entire framework can be trained in an end-to-end manner by minimizing the negative log-likelihood of the predicted traffic flow-sequence with the ground truth in training data. The only difference compared with the training of standard neural network is the forward propagation of the DE-Net part needs to calculate using the neural ODE solver in Eq.~\eqref{eq:traj-z}.
% (more details in Appendix).
% ~\cite{chen2018neural}.

{\em Remark:} In our model, the physical generation process of traffic flow in a potential energy field is expressed as deep residual GCN based DE-Net, providing a good explanation for our model structure. In other words, our model is not a complete ``black-box'' as other deep learning models. It could be considered as a kind of physics-guided generative model and therefore be expected to have better performance. Moreover, the elaborate residual DE-Net structure and the measure of uncertainty in potential energy fields also have the potential to improve the prediction performance. % of our model.

%Our model decouples the dynamics of the hidden traffic system, the encoding process of potential energy fields, and the likelihood of traffic flow observations so that each of them can be specified on its own. Meanwhile, the posterior of potential energy fields provides an measure of uncertainty, which is not available in standard sequence-to-sequence models.

\section{Experiments}\label{sec:expt}

\begin{table*}[t]\small
  \centering
%   \resizebox{\textwidth}{!}{
    \begin{tabular}{ccc|cccccccccc|c}
    \toprule
    \multicolumn{1}{c||}{} & \multicolumn{1}{c|}{T} & Metric & HA    & VAR   & GRU   & STGCN & DCRNN & GWNET & AGCRN & MTGNN & LODE & OLSTM & STDEN \\
    \hline
    \midrule
    \multicolumn{1}{c||}{\multirow{9}[6]{*}{\rotatebox{90}{GT-221}}} & \multicolumn{1}{c|}{\multirow{3}[2]{*}{\rotatebox{90}{15 min}}} & MAE   & 1.324  & 1.125  & 0.988  & 0.976  & 0.976  & 0.967  & 0.968  & 0.962  & 0.981  & 0.978  & \textbf{0.865 } \\
    \multicolumn{1}{c||}{} & \multicolumn{1}{c|}{} & RMSE  & 1.835  & 1.581  & 1.511  & 1.528  & 1.520  & 1.516  & 1.502  & 1.505  & 1.531  & 1.522  & \textbf{1.317 } \\
    \multicolumn{1}{c||}{} & \multicolumn{1}{c|}{} & MAPE  & 61.47 & 52.34 & 42.77 & 41.40 & 41.10 & 40.70 & 41.18 & 40.88 & 42.11 & 42.87 & \textbf{36.72} \\
    \cline{2-14}
    \multicolumn{1}{c||}{} & \multicolumn{1}{c|}{\multirow{3}[2]{*}{\rotatebox{90}{30 min}}} & MAE   & 1.324  & 1.141  & 1.000  & 0.988  & 0.994  & 0.980  & 0.981  & 0.976  & 0.992  & 0.986  & \textbf{0.872 } \\
    \multicolumn{1}{c||}{} & \multicolumn{1}{c|}{} & RMSE  & 1.835  & 1.620  & 1.541  & 1.552  & 1.553  & 1.540  & 1.525  & 1.531  & 1.544  & 1.561  & \textbf{1.328 } \\
    \multicolumn{1}{c||}{} & \multicolumn{1}{c|}{} & MAPE  & 61.47 & 52.03 & 42.95 & 42.08 & 41.75 & 41.23 & 41.73 & 41.40 & 42.71 & 41.80 & \textbf{37.37} \\
    \cline{2-14}    
    \multicolumn{1}{c||}{} & \multicolumn{1}{c|}{\multirow{3}[2]{*}{\rotatebox{90}{1 hour}}} & MAE   & 1.324  & 1.164  & 1.018  & 1.008  & 1.020  & 1.006  & 1.004  & 1.003  & 1.011  & 1.007  & \textbf{0.896 } \\
    \multicolumn{1}{c||}{} & \multicolumn{1}{c|}{} & RMSE  & 1.835  & 1.835  & 1.578  & 1.585  & 1.601  & 1.581  & 1.561  & 1.576  & 1.588  & 1.579  & \textbf{1.360 } \\
    \multicolumn{1}{c||}{} & \multicolumn{1}{c|}{} & MAPE  & 61.47 & 51.51 & 43.29 & 43.20 & 42.46 & 42.12 & 42.79 & 42.36 & 43.28 & 43.22 & \textbf{39.04} \\
    \midrule
    \midrule
    \multicolumn{1}{c||}{\multirow{9}[6]{*}{\rotatebox{90}{WRS-393}}} & \multicolumn{1}{c|}{\multirow{3}[2]{*}{\rotatebox{90}{15 min}}} & MAE   & 1.239  & 1.070  & 0.823  & 0.803  & 0.803  & 0.801  & 0.799  & 0.792  & 0.818  & 0.809  & \textbf{0.730 } \\
    \multicolumn{1}{c||}{} & \multicolumn{1}{c|}{} & RMSE  & 1.735  & 1.509  & 1.393  & 1.400  & 1.386  & 1.390  & 1.382  & 1.376  & 1.402  & 1.391  & \textbf{1.241 } \\
    \multicolumn{1}{c||}{} & \multicolumn{1}{c|}{} & MAPE  & 64.08 & 55.51 & 35.52 & 33.20 & 33.51 & 33.50 & 33.28 & 32.91 & 35.33 & 35.28 & \textbf{31.62} \\
    \cline{2-14}
    \multicolumn{1}{c||}{} & \multicolumn{1}{c|}{\multirow{3}[2]{*}{\rotatebox{90}{30 min}}} & MAE   & 1.239  & 1.097  & 0.837  & 0.818  & 0.826  & 0.818  & 0.814  & 0.807  & 0.827  & 0.818  & \textbf{0.737 } \\
    \multicolumn{1}{c||}{} & \multicolumn{1}{c|}{} & RMSE  & 1.735  & 1.568  & 1.429  & 1.436  & 1.435  & 1.427  & 1.414  & 1.411  & 1.433  & 1.428  & \textbf{1.244 } \\
    \multicolumn{1}{c||}{} & \multicolumn{1}{c|}{} & MAPE  & 64.08 & 55.72 & 36.00 & 33.69 & 34.46 & 33.95 & 33.96 & 33.34 & 35.77 & 35.12 & \textbf{32.04} \\
    \cline{2-14}
    \multicolumn{1}{c||}{} & \multicolumn{1}{c|}{\multirow{3}[2]{*}{\rotatebox{90}{1 hour}}} & MAE   & 1.239  & 1.139  & 0.856  & 0.841  & 0.858  & 0.845  & 0.841  & 0.834  & 0.853  & 0.848  & \textbf{0.745 } \\
    \multicolumn{1}{c||}{} & \multicolumn{1}{c|}{} & RMSE  & 1.735  & 1.653  & 1.477  & 1.490  & 1.500  & 1.488  & 1.468  & 1.469  & 1.482  & 1.478  & \textbf{1.266 } \\
    \multicolumn{1}{c||}{} & \multicolumn{1}{c|}{} & MAPE  & 64.08 & 56.01 & 36.10 & 34.31 & 35.83 & 34.37 & 34.93 & 34.01 & 35.87 & 35.14 & \textbf{33.25} \\
    \midrule
    \midrule
    \multicolumn{1}{c||}{\multirow{9}[6]{*}{\rotatebox{90}{ZGC-564}}} & \multicolumn{1}{c|}{\multirow{3}[2]{*}{\rotatebox{90}{15 min}}} & MAE   & 1.120  & 0.978  & 0.729  & 0.714  & 0.715  & 0.716  & 0.709  & 0.708  & 0.723  & 0.718  & \textbf{0.623 } \\
    \multicolumn{1}{c||}{} & \multicolumn{1}{c|}{} & RMSE  & 1.522  & 1.402  & 1.229  & 1.218  & 1.230  & 1.232  & 1.210  & 1.207  & 1.226  & 1.219  & \textbf{1.026 } \\
    \multicolumn{1}{c||}{} & \multicolumn{1}{c|}{} & MAPE  & 62.68 & 53.72 & 33.82 & 32.63 & 31.99 & 31.31 & 32.19 & 32.64 & 32.98 & 32.75 & \textbf{28.81} \\
    \cline{2-14}
    \multicolumn{1}{c||}{} & \multicolumn{1}{c|}{\multirow{3}[2]{*}{\rotatebox{90}{30 min}}} & MAE   & 1.120  & 0.991  & 0.734  & 0.724  & 0.727  & 0.725  & 0.717  & 0.714  & 0.730  & 0.725  & \textbf{0.628 } \\
    \multicolumn{1}{c||}{} & \multicolumn{1}{c|}{} & RMSE  & 1.522  & 1.393  & 1.242  & 1.241  & 1.258  & 1.255  & 1.228  & 1.222  & 1.263  & 1.252  & \textbf{1.036 } \\
    \multicolumn{1}{c||}{} & \multicolumn{1}{c|}{} & MAPE  & 62.68 & 53.64 & 33.99 & 33.13 & 32.34 & 31.32 & 32.66 & 32.90 & 33.78 & 32.65 & \textbf{29.14} \\
    \cline{2-14}
    \multicolumn{1}{c||}{} & \multicolumn{1}{c|}{\multirow{3}[2]{*}{\rotatebox{90}{1 hour}}} & MAE   & 1.120  & 1.019  & 0.747  & 0.738  & 0.746  & 0.743  & 0.733  & 0.731  & 0.742  & 0.740  & \textbf{0.657 } \\
    \multicolumn{1}{c||}{} & \multicolumn{1}{c|}{} & RMSE  & 1.522  & 1.441  & 1.266  & 1.272  & 1.298  & 1.296  & 1.259  & 1.255  & 1.281  & 1.274  & \textbf{1.033 } \\
    \multicolumn{1}{c||}{} & \multicolumn{1}{c|}{} & MAPE  & 62.68 & 53.85 & 34.55 & 33.78 & 32.86 & 32.50 & 33.24 & 33.54 & 33.98 & 33.36 & \textbf{31.58} \\
    % \midrule
    % \multicolumn{3}{c|}{Average Rank} & 11    & 10    & 9     & 5     & 7     & 4     & 3     & 2     & 8     & 6     & 1 \\
    \bottomrule
    \end{tabular}%
    % }
  \caption{Model comparison on metrics MAE/RMSE/MAPE, where MAPE is in \%. Our \name \emph{significantly} outperforms all competing baselines with regard to all metrics over all datasets according to Student's $t$-test at level 0.01.}\label{tab:h_cmp}%
\end{table*}%

\subsection{Datasets}

We evaluate the performance of our model over the real-world urban traffic dataset collected by the Beijing Municipal Commission of Transport, which contains trajectories of 40,000 taxies in Beijing from April 1st 2015 to July 31st 2015 (totally 4 months). These trajectories mapped to the road networks using the map matching algorithm. We statistic the traffic flow by counting the number of taxis on each road segment during every 5-minute time interval, resulting in 288 data points per day.
% ~\cite{can2018fast}

Due to Beijing road networks are too complex, we select three sub-networks to construct datasets of our experiments. Without loss of dataset diversity, the selected sub-networks have different urban functions. The first one is a well-known entertainment area, \ie Gong Ti (Workers' Stadium), which has 221 road segments, the second is the area around Beijing West Railway Station, which has 393 road segments, and the last area is around the biggest business park in Beijing, \ie Zhongguancun (China Silicon Valley), which has 564 road segments. We denote the three datasets as GT-221, WRS-393, and ZGC-564 respectively. 
% The statistics of the three datasets are provided in Appendix \ref{asec:data}.
We split each dataset into the training, validation, and test sets with a ratio of 7:1:2. For multi-step prediction, we use one-hour historical traffic flow data (12 time steps) to predict the next hour's.

\subsection{Experimental Settings}

\paratitle{Baselines.} We consider ten baselines that belong to three classes. (1) \emph{Time series modeling:} We take Historical Average (HA), Vector Auto-Regression (VAR) and GRU as baselines. The history traffic flow-sequence are treated as purely time series to predict the futures state without consideration of spatial information. (2) \emph{Graph-based spatio-temporal methods:} The classical graph-based tarffic prediction methods such as Diffusion Convolution Recurrent Neural Network (DCRNN)~\cite{li2018diffusion}, Spatial-Temporal Graph Convolutional Network (STGCN)~\cite{yu2018spatio} and Graph WaveNet (GWNET)~\cite{wu2019graph}, and state-of-the-art models such as Adaptive Graph Convolutional Recurrent Network (AGCRN)~\cite{bai2020adaptive} and MTGNN~\cite{wu2020connecting} are used for comparison. (3) \emph{Approaches based on neural differential equation networks}: Here we use two classical methods Latent-ODE (LODE)~\cite{rubanova2019latent} and ODE-LSTM (OLSTM)~\cite{lechner2020learning} for comparison. Latent-ODE generalizes RNNs to have continuous-time hidden dynamics defined by ordinary differential equations (ODEs). ODE-LSTM is a novel long short term memory network, that possesses a continuous-time output state, and consequently modifies its internal dynamical flow to a continuous-time model.

\paratitle{Settings.} The settings of \name contains the following two parts:
(1) Settings of the DE-Net part. We model the dynamics of potential energy in latent space using an adaptive method $dopri5$~\cite{dormand1980family}, and conduct grid search on the latent dimension over $\{1, 2, 4, 8\}$.
(2) Settings of the encoder. GRU is used to encode the distribution of the initial value of the PEF-sequence. The number of hidden units in GRU is searched over $\{16, 32, 64, 128\}$.
More implementation details about our \name and other baselines settings are given in Appendix. We conduct experiments of all deep learning models with 7 different seeds and report the mean results.

\subsection{Performance Comparison}

\tableautorefname{~\ref{tab:h_cmp}} shows the comparison of different approaches for 15 minutes, 30 minutes, and 1 hour ahead forecasting on three datasets. These methods are evaluated by three commonly used metrics in traffic flow prediction, including mean absolute error (MAE), root mean square error (RMSE), and mean absolute percentage error (MAPE). 

There are four observations from \tableautorefname{~\ref{tab:h_cmp}}. (1) The graph neural network-based spatio-temporal methods generally outperform the time series models, which emphasizes the importance of modeling the spatial correlations of road networks for traffic prediction. (2) Our \name improves spatio-temporal methods with a significant margin (an average 10.29\%/13.49\%/6.03\% improvement on MAE/RMSE/MAPE compared with the second best method) and achieves the best performance for all horizons on all the metrics, which shows the effectiveness of the potential energy field DE to model the continuous spatial and temporal dynamics. (3) The approaches based on purely neural differential equation network are not effective than the spatio-temporal ones. Because they ignore the spatial correlation which is important to traffic flow prediction. However, they outperform the traditional time series methods. (4) Traditional methods including HA and VAR are not good enough due to their incapability of handling complex and non-linear spatio-temporal data. Besides, the performance of HA is invariant since this method does not depend on short-term data.

\subsection{Ablation Study}

To better illustrate the effectiveness of the {\em Potential Energy Field DE} in Eq.~\eqref{eq:potn_fin}, we compare \name with the following variants: (1) \uem{UnkP}, which means unknown physics and replaces the potential energy field DE with a fully connected neural network. (2) \uem{IncP}, which denotes incomplete physics and ignores the energy volume factor $\bm{\phi}$ in the potential energy field DE, making it incomplete. 

\begin{figure}[t]
  \centering
  \includegraphics[width=0.95\columnwidth]{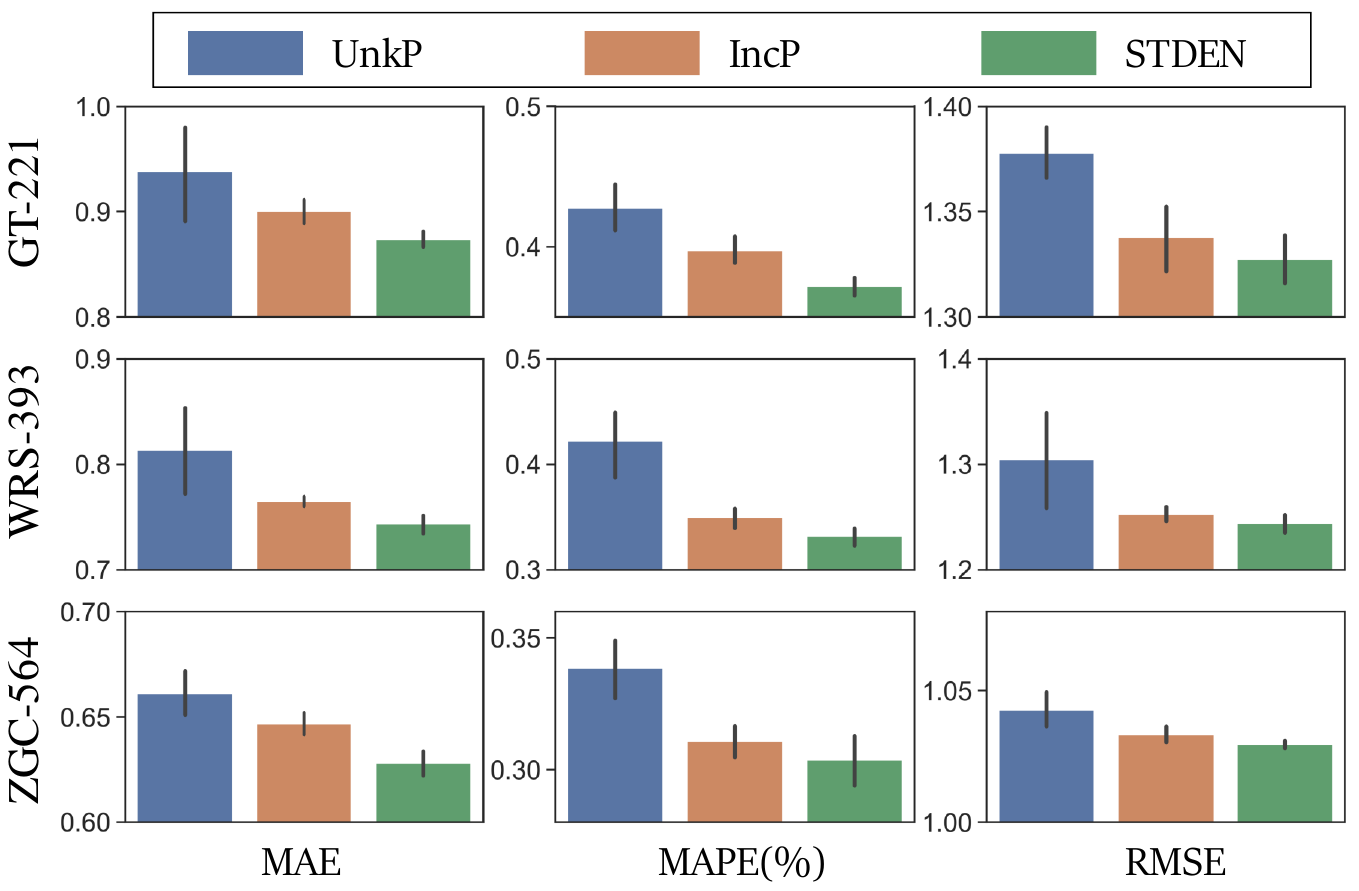}
  \vspace{-.1cm}
  \caption{Evaluation on potential energy field differential equation over all datasets.}\label{fig:ab}
  \vspace{-.3cm}
\end{figure}

\figureautorefname{~\ref{fig:ab}} shows the comparison of the two variants with regard to all metrics for different datasets. From the results, we have the following observations: (1) \name surpasses \uem{NoP} by a large margin, demonstrating the superiority of deploying potential energy fields to model traffic flow. (2) That \uem{IncP} beats \uem{UnkP} shows the effectiveness of the potential energy field DE. The guidance of physical process expressed by the potential energy field DE introduces the diffusion process of energy in space. This allows \uem{IncP} to capture the changing trend of potential energy fields throughout the road network. (3) However, \uem{IncP} performs slightly worse than \name because of the absence of the energy volume factor $\bm{\phi}$. Without $\bm{\phi}$,  \uem{IncP} has to infer the potential energy only through the energy density, which is as hard as to calculate the mass of an object from density without object volume.

\begin{figure}[t]
  \centering
  \subfigure[Epochs vs. NFE (GT-221).]{
    \includegraphics[width=0.3\columnwidth]{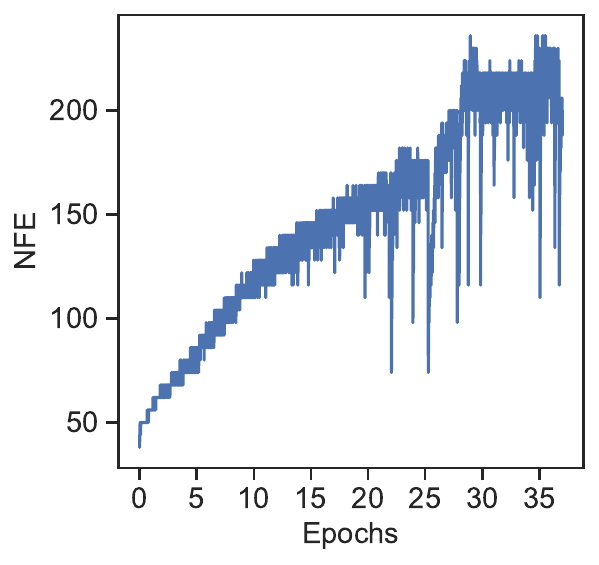}\label{fig:gm_nfe}
  }
  \subfigure[Epochs vs. NFE (WRS-393).]{
    \includegraphics[width=0.3\columnwidth]{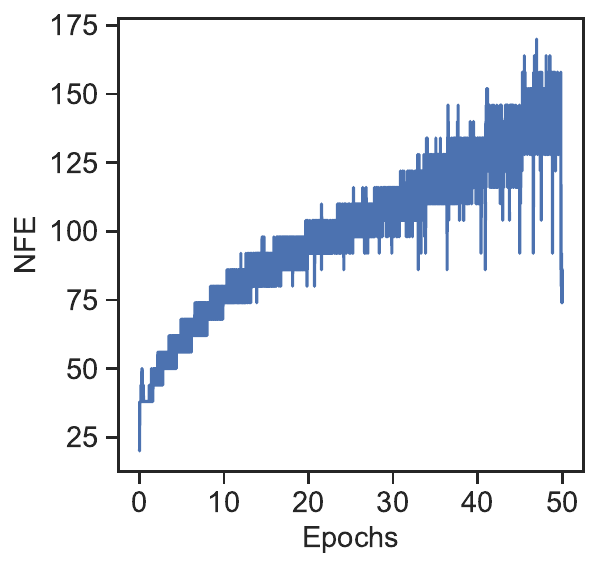}\label{fig:xz_nfe}
  }
  \subfigure[Epochs vs. NFE (ZGC-564).]{
    \includegraphics[width=0.3\columnwidth]{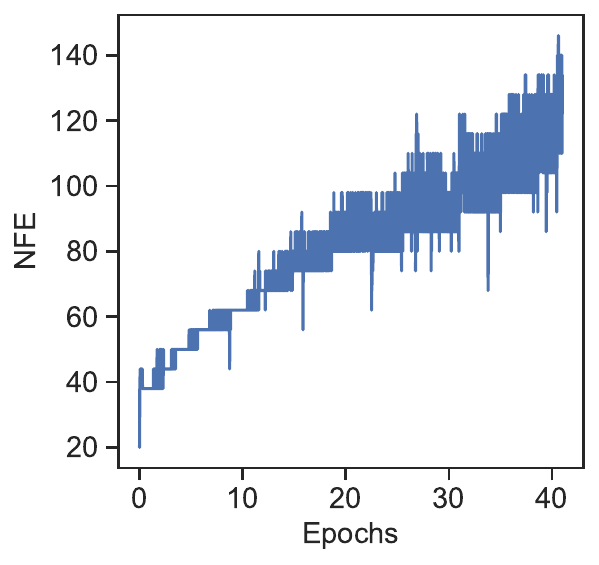}\label{fig:rm_nfe}
  }
  \subfigure[NFE vs. MAE (GT-221).]{
    \includegraphics[width=0.3\columnwidth]{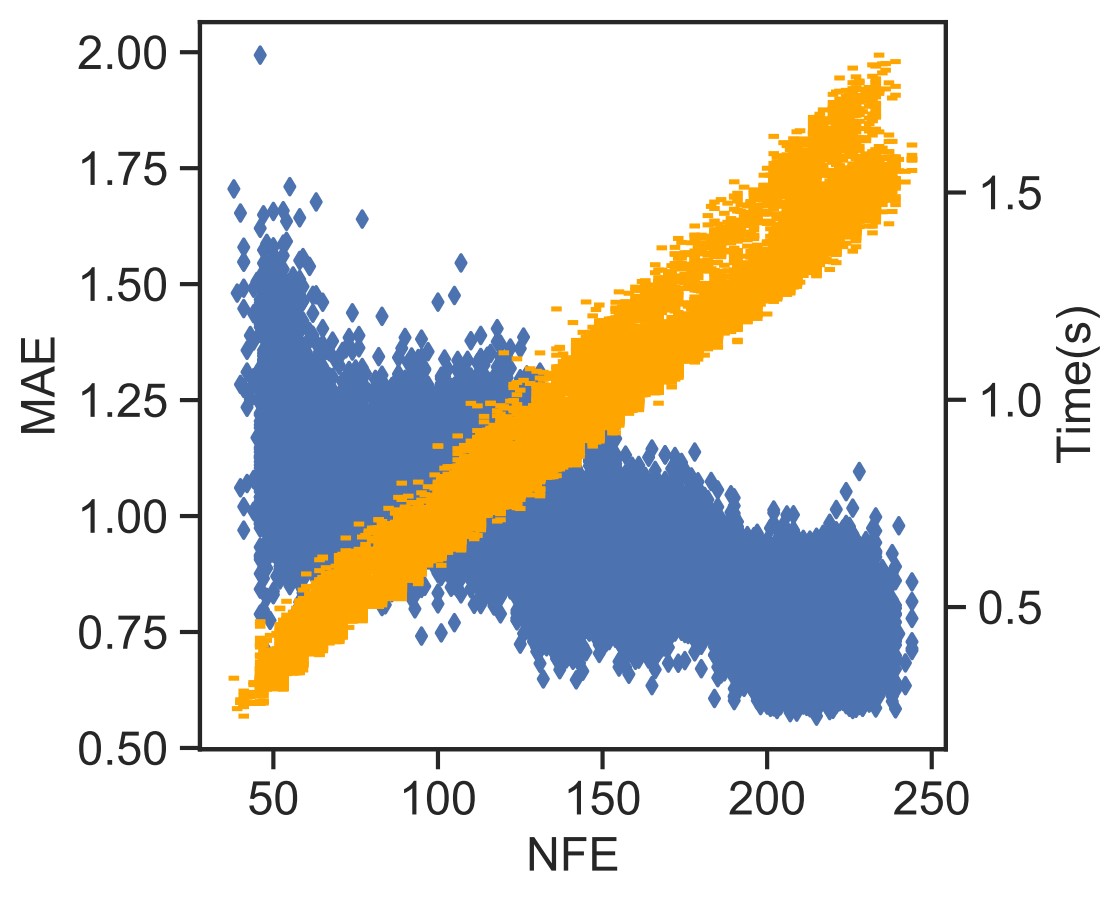}\label{fig:gm_mae}
  }
  \subfigure[NFE vs. MAE (WRS-393).]{
    \includegraphics[width=0.3\columnwidth]{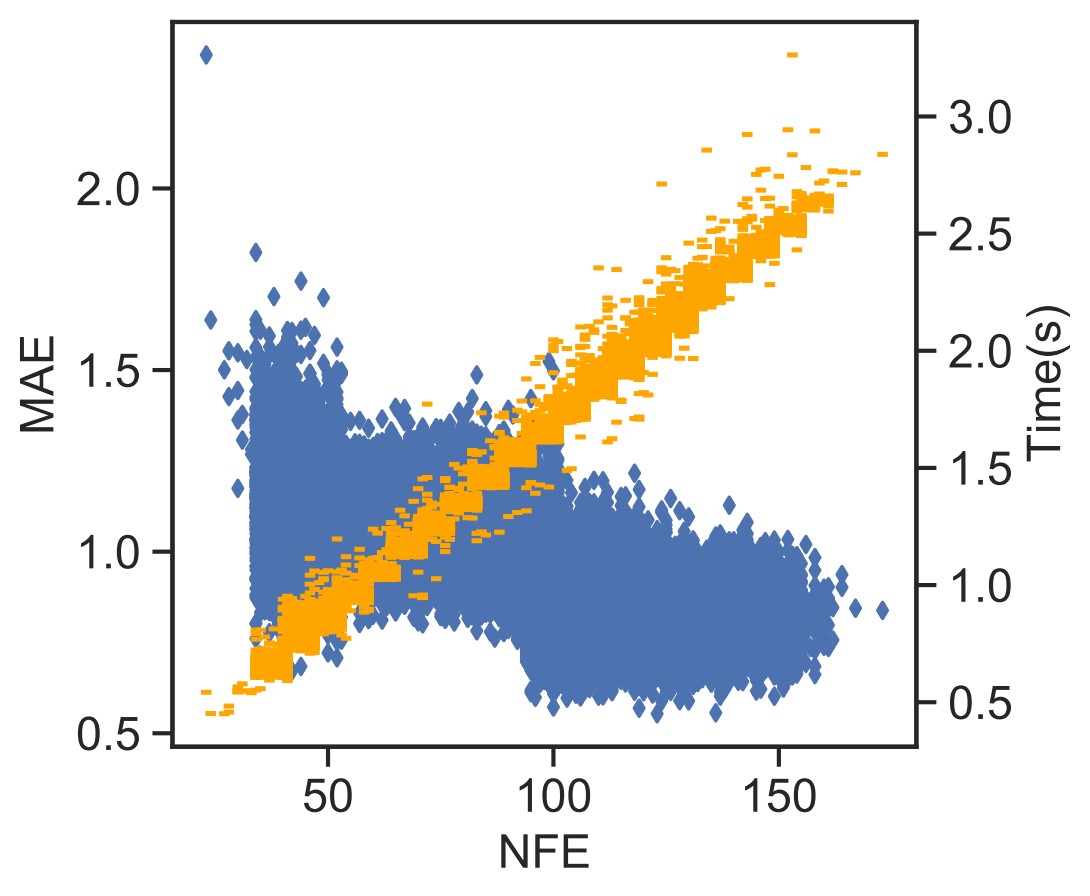}\label{fig:xz_mae}
  }
  \subfigure[NFE vs. MAE (ZGC-564).]{
    \includegraphics[width=0.3\columnwidth]{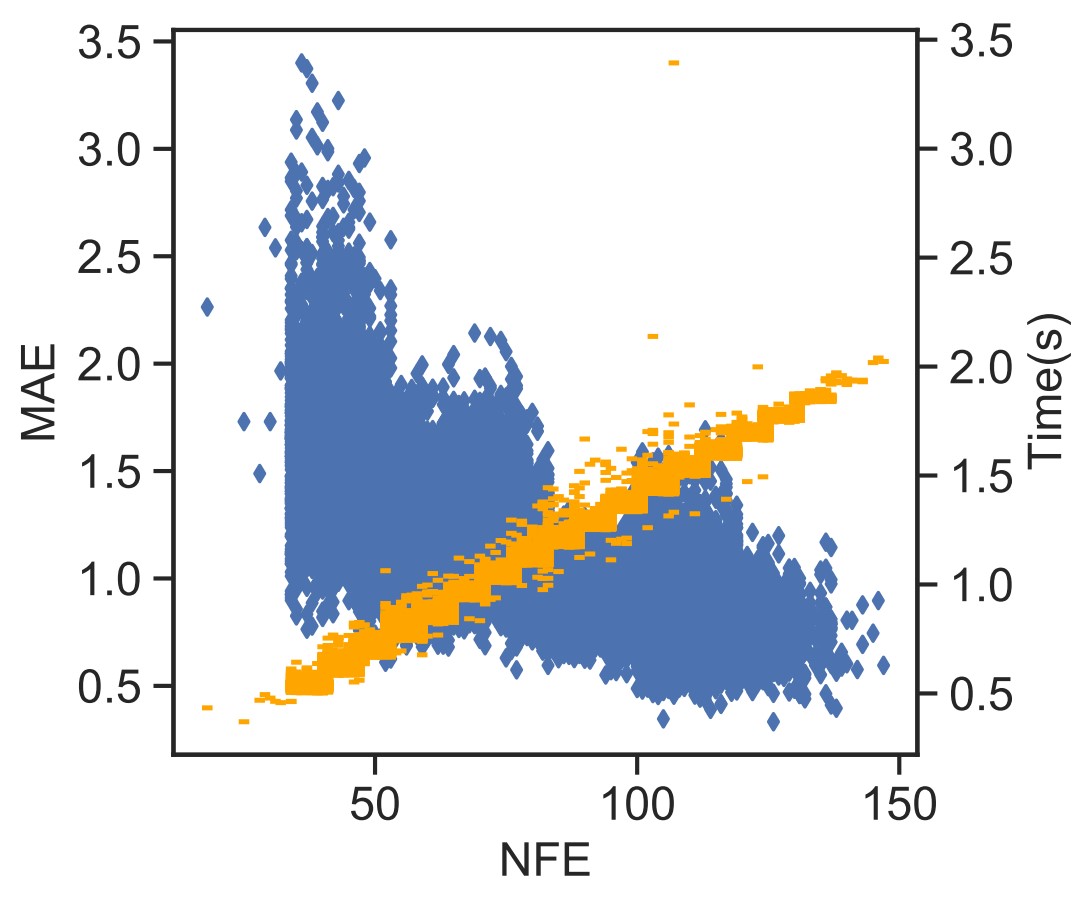}\label{fig:rm_mae}
  }
  %\vspace{-.3cm}
  \caption{Computation overhead vs. Prediction accuracy. NFE denotes number of function evaluation.}\label{fig:comp}
  %\vspace{-.3cm}
\end{figure}

\subsection{Computation Cost}

During the solving process of the differential equation network, we utilize an ODE solver which allows our method to dynamically balance the trade-off between prediction accuracy and the computation cost. As shown in \figureautorefname~\ref{fig:gm_nfe}, \ref{fig:xz_nfe} and \ref{fig:rm_nfe}, required number of function evaluation (NFE) increases consistently with the epochs, there is no stable stage because we set early stopping for the training process. Note that NFE varies with the difficulty of tasks. For example, flow prediction of GT-221 is the most complicated among all three tasks because the traffic flow in GT-221 changes more drastically. Therefore, the NFE is around 80\% more than that of ZGC-564. From \figureautorefname~\ref{fig:gm_mae}, \ref{fig:xz_mae} and \ref{fig:rm_mae}, it can be observed that the more evaluation (NFE), the lower the prediction error our method can achieve. This allows us to trade accuracy for faster response for emergency events during inference phase, which is very valuable in practice.

\subsection{Case Study}

\begin{figure}[t]
  \centering
  \includegraphics[width=0.98\columnwidth]{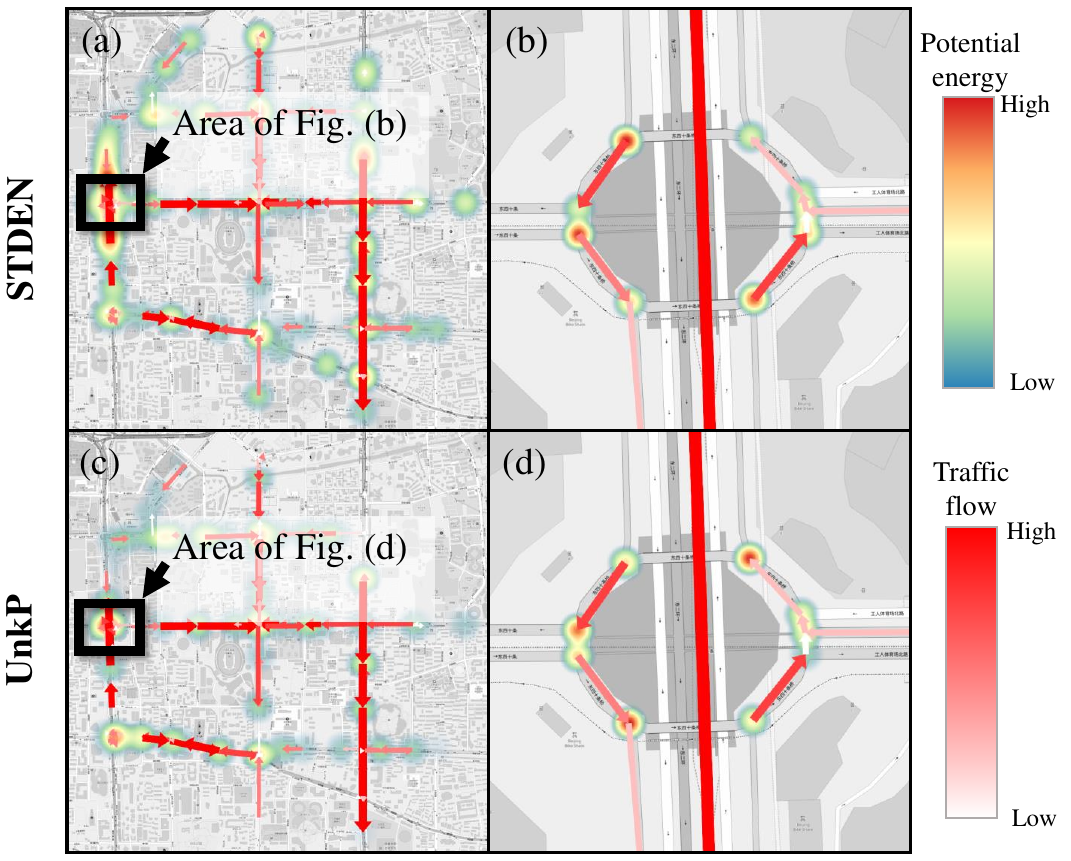}
  \caption{Visualization of the learned potential energy fields and real traffic flow on GT-221 dataset. The heat map represents the potential energy fields, while the arrows denote traffic flow with its volume reflected by the color and the arrow size. The potential energy fields learned by \name can interpret the traffic flow.}%\vspace{-.3cm}
  \label{fig:cs_flow_heat}%\vspace{-.3cm}
\end{figure}

In this part, we qualitatively analyze why our \name can yield good performance. To this end, we visualize the learned potential energy and the real traffic flow in \figureautorefname{~\ref{fig:cs_flow_heat}}.

\figureautorefname{~\ref{fig:cs_flow_heat}}(a) and (c) shows the potential energy learned by \name and UnkP by a heat map, with real traffic flows at the corresponding moment represented by arrows. To see more clearly how the potential energy drives traffic flow, we select loop areas with complex traffic for detailed analysis. In \figureautorefname{~\ref{fig:cs_flow_heat}}(b), traffic always flows along the road network from a place with high potential energy to a place with low potential energy. Besides, the gradient of potential energy roughly reflects the relative magnitude of the traffic flow. This shows that potential energy governed by physical equation captures the mechanism of traffic flow and can be interpreted as a force that drives traffic flow. However, without the guidance of potential energy field DE, the energy learned by neural networks in UnkP (\figureautorefname{~\ref{fig:cs_flow_heat}}(d)) has no physical meaning and can not indicate the traffic flow mechanism. For example, there are some traffic flows that transport from a low-value node to a high-value node in \figureautorefname{~\ref{fig:cs_flow_heat}}(d), which violates the physical constraint of the energy transport process.

In a word, the results in this section verifies that \name can capture the physical mechanism of traffic flow (\figureautorefname{~\ref{fig:cs_flow_heat}}) and generate accurate predictions (\tableautorefname{~\ref{tab:h_cmp}}) under the guidance of the potential energy field DE. This is because we combine the advantages of physics-based and data-driven methods.

\section{Related Work}\label{sec:relt}

\paratitle{Traffic Prediction.} Traffic prediction problem has been studied for decades, and existing methods mainly fall into two categories: \emph{physics-based} and \emph{data-driven}. In the former one, researchers apply different branches of traffic flow theory depending on the application problems~\cite{ni2015traffic}, such as the kinematic wave theory~\cite{daganzo2005moving}, the car-following theory~\cite{olstam2004comparison}, and the queuing theory~\cite{cascetta2013transportation}. However, they may only capture limited dynamics of real-world traffic, resulting in low-quality estimation of traffic flow. The \emph{data-driven} approaches have drawn considerable attentions~\cite{xie2020urban}. The shallow machine learning methods for traffic prediction usually base on the stationary assumptions (\eg ARIMA and Kalman filtering\cite{lippi2013short}), leading to limited representation power. Deep learning methods are free from stationary assumptions and effective to capture complex non-linearity using models such as recurrent neural networks~\cite{li2018diffusion} and temporal convolutional networks~\cite{wu2019graph, li2021spatial}. Because traffic data is spatial correlated, CNN~\cite{tang2020joint} and its extension to arbitrary graphs~\cite{song2020spatial, tian2021spatial} are utilized to capture spatial correlations. Although temporal dependencies and spatial correlations have been considered in these methods, the lack of physical knowledge leads to a lack of generalization ability to out-of-sample scenarios.

\paratitle{Physics-Guided Deep Learning.} Many recent studies have proposed to integrate physics-based modeling approaches with state-of-the-art deep learning techniques, giving birth to a field called ``Physics-Guided Deep Learning''. One can introduce additional physics-based penalty in loss function of neural networks~\cite{shi2021physicsinformed}. Some efforts also lie in combining physics-based models with deep learning. For example, \cite{wang2020towards} presents a hybrid framework that combines turbulent flow simulation with deep learning. \cite{ji2020interpret} introduces the physical potential energy field concept into deep learning to achieve grid-based urban traffic prediction while preserving community information~\cite{wang2017community}. However physics-based models governed by differential equations are usually continuous in time, while deep learning are dominated by discrete models~\cite{ruthotto2019deep}. Recent advances propose a treatment of neural networks equipped with a continuum of layers~\cite{chen2018neural}, allowing a more accurate and natural modeling of physical principles in real world, \eg hydropower generation~\cite{zhou2020forecasting} and reservoir flow~\cite{zhou2021forecasting}. To the best of our knowledge, we are the first to introduce physics-guided deep learning into road network-based traffic flow prediction.

\section{Conclusion and Future Work}

In this paper, we introduced potential energy fields as the dominant force to drive traffic flows and derived an differential equation to describe the physical mechanism of the traffic potential energy fields. Based on the differential equations, we proposed a novel Spatio-Temporal Differential Equation Network (\name), blending deep learning into physical mechanism modeling, for physics-guided traffic flow prediction. Extensive experiments on three real-world traffic datasets demonstrated the effectiveness of the proposed \name. A case study further verified that our model can capture the mechanism of urban traffic and generate accurate predictions with physical meaning. However, we have noticed that the major efficiency bottleneck of our model is the evolution modeling of potential energy fields. In the future, we plan to reduce the number of function evaluations in ODE solver while preserving the model performance, and explore \name on more datasets.

\section*{Acknowledgments}

The work of J. Ji, J. Wang, J. Jiang and H. Zhang was supported by the National Key R\&D Program of China (2019YFB2101804), the National Natural Science Foundation of China (Grant No. 72171013, 82161148011, 92046010), CCF-DiDi Gaia Collaborative Research Funds for Young Scholars, and the Fundamental Research Funds for the Central Universities (Grant No. YWF-21-BJ-J-839).

% \clearpage
\bibliography{base}

\end{document}